# THG: Transformer with Hyperbolic Geometry

Zhe Liu
School of Computer Science and Technology
Dalian University of Technology
Dalian, China
njjnlz@mail.dlut.edu.cn

Yibin Xu
School of Computer Science and Technology
Dalian University of Technology
Dalian, China
19xyb@ mail.dlut.edu.cn

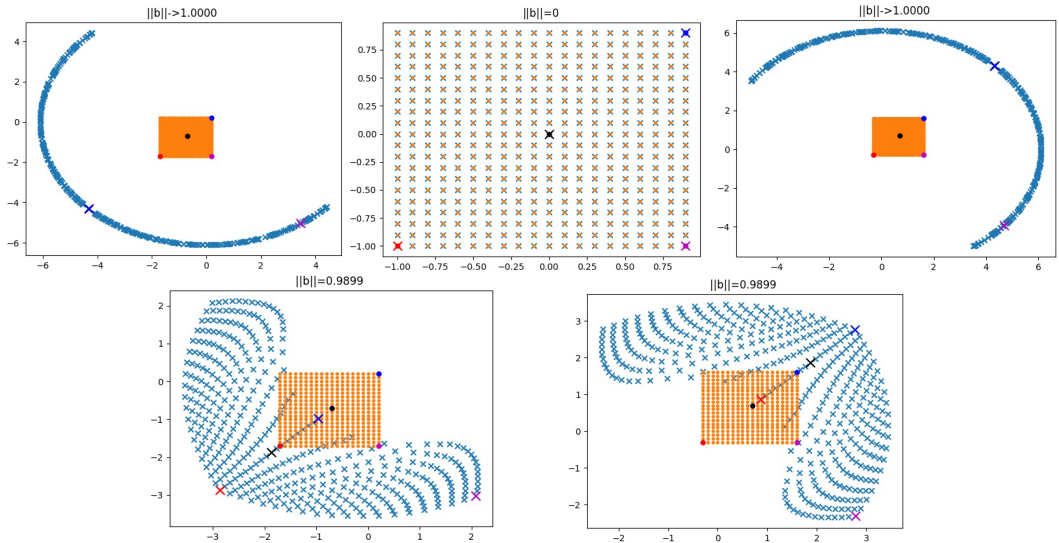

Figure 1: Five examples of 2-dimensional Linear, where the orange dots and blue crosses represent the output points of the Liner in Euclidean space and Hyperbolic space (i.e. Poincaré ball), respectively. For better understanding, we emphasis eight points, that the different outputs points with the same input are in the other same color. All dimensions of b in the left two are smaller than 0, and that in the right two are

## Abstract

Transformer model architectures have become an indispensable staple in deep learning lately for their effectiveness across a range of tasks. Recently, a surge of "X-former" models have been proposed which improve upon the original Transformer architecture. However, most of these variants make changes only around the quadratic time and memory complexity of self-attention, i.e. the dot product between the query and the key. What's more, they are calculate solely in Euclidean space. In this work, we propose a novel Transformer with Hyperbolic Geometry (THG) model, which take the advantage of both Euclidean space and Hyperbolic space. THG makes improvements in linear transformations of self-attention, which are applied on the input sequence to get the query and the key, with the proposed hyperbolic linear. Extensive experiments on sequence labeling task, machine reading comprehension task and classification task demonstrate the effectiveness and generalizability of our model. It also demonstrates THG could alleviate overfitting.

## 1 Introduction

Transformer models (Vaswani et al., 2017) are pervasive and have demonstrated impressive results on a variety of Natural Language Processing tasks (Devlin et al., 2019; Lee et al., 2020), such as sequence labeling, machine reading comprehension and classification.

The key innovation in Transformers is the introduction of a self-attention mechanism. A well-known concern with self-attention is quadratic time and memory complexity (Tay et al., 2020). Therefore such a surge of Transformer model variants have been proposed recently to address this problem (Wang et al., 2020; Choromanski et al., 2020; Zaheer et al., 2020). However, there are other areas that can be improved. We find that most of the existing Transformer model are designed solely in Euclidean space.

There also has been a recent trend to embed networks into hyperbolic space (Gulcehre et al., 2018; Chami et al., 2020), because hyperbolic geometry can naturally reflect some properties of complex networks (Krioukov et al., 2010). One key property of Hyperbolic spaces is that they expand exponentially, while Euclidean spaces expand polynomially. However, these methods get better performance only in low dimensions but achieve worse results in high dimensions, while using high dimensions are recent trend (Devlin et al., 2019). Moreover, they leverage hyperbolic distance which is time and memory consuming and is a distance-preserving operation.

In this paper, we propose a novel **T**ransformer with **H**yperbolic **G**eometry (THG) model to take the advantage of both Euclidean space and Hyperbolic space. Specifically, we propose Hyperbolic Linear to generate hyperbolic query and hyperbolic key. For effectiveness and time-efficient, we use dot product in Euclidean space rather than hyperbolic distance. In summary, the major contributions can be summarized as follows:

- We propose THG to take the advantage of both Euclidean space and Hyperbolic space.
- We propose Hyperbolic Linear to generate hyperbolic query and hyperbolic key, of which the distribution can automatically fit the distribution of the target task through back propagation. The distribution makes that THG could alleviate overfitting.
- We conduct extensive experiments to evaluate the performance of THG on sequence labeling, machine reading comprehension and classification task. The results show the effectiveness and generalizability of THG.

## 2 Methodology

Firstly, to facilitate the understanding, we give some necessary background on hyperbolic geometry (Section 2.1). Secondly, to leverage the properties of hyperbolic space, we propose hyperbolic linear (Section 2.2). Finally, we present THG shown in Figure 2. (Section 2.3)

### 2.1 Hyperbolic Geometry

We introduce some key notions in hyperbolic geometry briefly. In this work, we adopt the $d$-dimensional Poincaré ball model with negative curvature $-c$ $(c > 0)$ : $\mathrm{B}^{d,c} = \{\mathbf{x} \in \mathbb{R}^d \mid \|\mathbf{x}\|^2 < \frac{1}{c}\}$ , where $\|\cdot\|$ denotes the $L_2$ norm, due to its feasibility for gradient optimization (Balazevic et al., 2019).

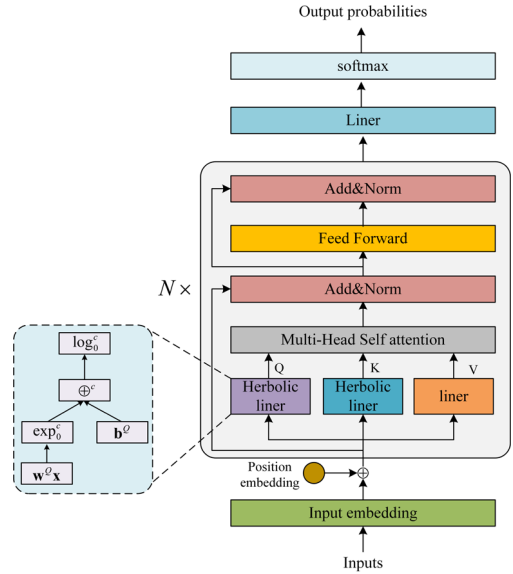

Figure 2: The framework of THG model.

**Projection:** For each point $\mathbf{x} \in \mathrm{B}^{d,c}$, there is a tangent space containing all possible directions of paths $\mathbf{v} \in \mathbb{R}^d$ in $\mathrm{B}^{d,c}$ leaving from $\mathbf{x}$ , the calculation in which could be done just like in the Euclidean space. Ganea et al (2018) propose the exponential map to project the feature from tangent space into the hyperbolic space and the logarithmic map to project it back:

$$\exp_{\mathbf{0}}^c(\mathbf{v}) = \tanh(\sqrt{c}\|\mathbf{v}\|)\frac{\mathbf{v}}{\sqrt{c}\|\mathbf{v}\|} \qquad (1)$$

$$\log_{\mathbf{0}}^c(\mathbf{x}) = \operatorname{arctanh}(\sqrt{c}\|\mathbf{x}\|)\frac{\mathbf{x}}{\sqrt{c}\|\mathbf{x}\|} \qquad (2)$$

Here we assume that the feature lies in the tangent spaces at the point $\mathbf{x} = \mathbf{0}$ .

**Mobius addition:** Euclidean addition is unsuitable in the hyperbolic space, because adding two points in the Poincaré ball might result in a point outside the ball. Therefore, there is a specialized addition operation for hyperbolic space called Mobius addition:

$$\mathbf{x} \oplus^c \mathbf{y} = \frac{(1 + 2c\mathbf{x}^T\mathbf{y} + c\|\mathbf{y}\|^2)\mathbf{x} + (1 - c\|\mathbf{x}\|^2)\mathbf{y}}{1 + 2c\mathbf{x}^T\mathbf{y} + c^2\|\mathbf{x}\|^2\|\mathbf{y}\|^2} \qquad (3)$$

**Hyperbolic distance:** Given two points $\mathbf{x}, \mathbf{y} \in \mathrm{B}^{d,c}$ , the hyperbolic distance is defined as follows:

$$d^c(\mathbf{x}, \mathbf{y}) = \frac{2}{\sqrt{c}} \operatorname{arctanh}(\sqrt{c} \, \| -\mathbf{x} \oplus^c \mathbf{y} \|) \qquad (4)$$

**Riemannian gradient:** While performing back-propagation, we adopt the Riemannian SGD algorithm (Bonnabel, 2013) to update the parameters in the hyperbolic space. The Riemannian gradient $\nabla_B$ is computed as follows:

$$\nabla_B = \frac{\left(1 - \|\theta\|^2\right)^2}{4} \nabla_E \qquad (5)$$

where $\theta$ is the trainable parameters, and $\nabla_E$ denotes the Euclidean gradient.

## 2.2 Hyperbolic Linear

Linear are generally used in Euclidean neural networks. Analogue of linear transformation in Euclidean space, we propose hyperbolic linear, which is defined as follows:

$$y = \log_0^c \left( \exp_0^c (\mathbf{w} \mathbf{x}) \oplus^c \mathbf{b} \right) \qquad (6)$$

where $\mathbf{w}$ and $\mathbf{b}$ are trainable parameters. In Euclidean Linear, $\mathbf{w}$ are initialized by using kaiming initializer (He et al., 2015), and $\mathbf{b}$ is initialized by uniform distribution. However, in Hyperbolic Linear, $\mathbf{w}$ are initialized by orthogonal distribution (Saxe et al., 2014), and $\mathbf{b}$ is initialized by zeros.

To help make sense of hyperbolic linear, some properties will be expounded. The Euclidean orthogonal transformation $\mathbf{w}$ preserves the boundary plane at infinity and is simultaneously a hyperbolic isometry, which is proven by Cannon et al. (1997). What's more, as shown in Figure 1, Hyperbolic linear transformation recover the Euclidean linear transformation when $\mathbf{b}$ goes to zero vector. And when $\|\mathbf{b}\|$ goes to 1, all the point are getting closer and closer to the Poincaré shell. Finally, for different $\mathbf{b}$, the corresponding output points in the Euclidean space can be gotten by translating. That is to say, the distance of two arbitrary points is invariant. However, in the Hyperbolic space, for different $\mathbf{b}$, the shape of the corresponding output points set are changeable, and the distant of the any two points is variant. This means output points have different distributions for different $\mathbf{b}$.

## 2.3 The THG Model

Since hyperbolic space cannot be isometrically embedded into Euclidean space, to endow different subsets of Euclidean space with a hyperbolic metric, we propose Transformer with Hyperbolic Geometry (THG) model using hyperbolic linear, as shown in Figure 2.

We leverage hyperbolic linear, described in the section 2.2, instead of the linear in Euclidean space, to generate hyperbolic query and hyperbolic key with shared curvature $c$. In this way, for different $\mathbf{b}$, hyperbolic query and hyperbolic key would have different distributions. The distribution can automatically fit the target task through back propagation. With the task-oriented distribution, the calculation result of the compatibility function of the query with the corresponding key would be easier to attend to task-oriented information and Considering effectivity and the computational efficiency, we use dot product rather than hyperbolic distance as the compatibility function. It is clearly that the calculations of hyperbolic distance need more computational resource than dot product. Furthermore, when using hyperbolic distance as compatibility function, the distant of two points is invariant under pullback on transition functions connecting Euclidean and Hyperbolic space, which is proven by Cannon et al. (1997). In this way, for different $\mathbf{b}$, the distribution is same.

## 3 Experiments

### 3.1 Dataset and Evaluation Metrics

**Dataset:** For sequence labeling task, experiments are performed on biomedical named entity recognition datasets, i.e. BC2GM gene (Smith et al., 2008), BC4CHEMD chemical (Krallinger et al., 2015), BC5CDR chemical (Li et al., 2016), BC5CDR disease (Li et al., 2016), JNLPBA gene (Kim et al., 2004), Species-800 species (Pafilis et al., 2013). We use the pre-processed datasets provided by Lee et al. (2020).

For machine reading comprehension task, experiments are made on common domain named entity recognition datasets, i.e. English ACE 2004 (Doddington et al., 2005), English ACE 2005 (Walker et al., 2006) and English CoNLL 2003 (Sang and Meulder, 2003). We use the pre-processed datasets provided by Li et al. (2020).

For classification task, experiments are performed on relation extraction dataset, i.e. DocRED (Yao et al., 2019). We use the pre-processed datasets provided by Yao et al. (2019).

**Evaluation Metrics:** For biomedical NER evaluation, we use entity level $F$-score. For common domain NER evaluation, we use span level micro-averaged $F$-score. For RE, the evaluation on test set is done through CondaLab[1] and $F$-score is used in our experiments.

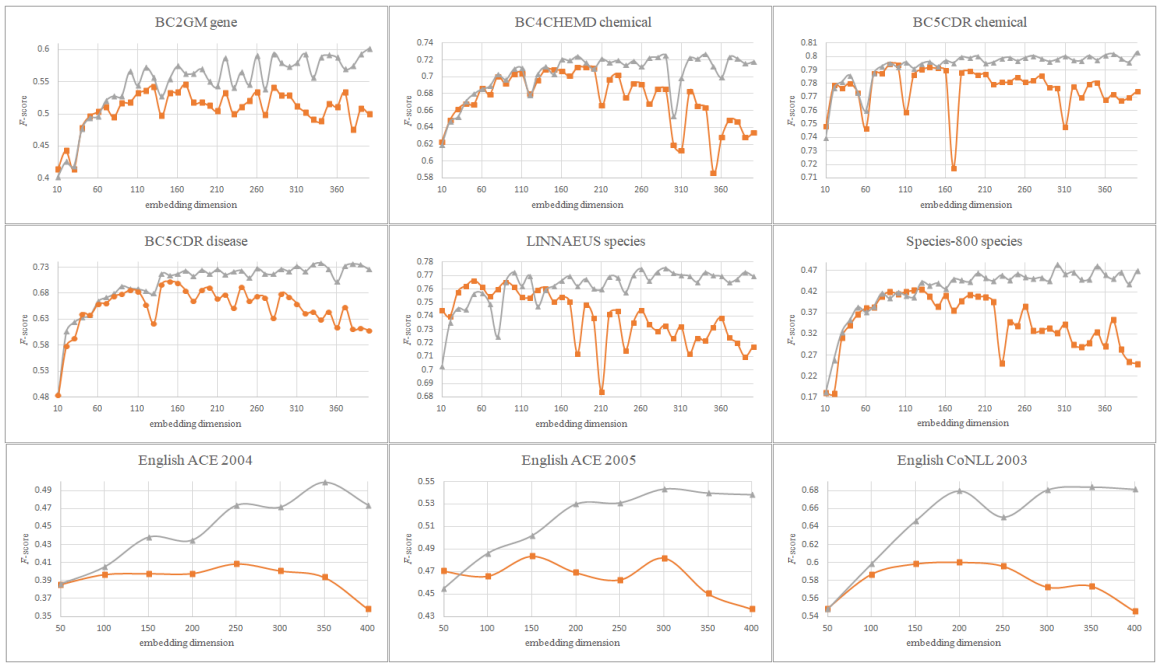

Figure 3: Results of sequence labeling tasks (the above six figures) and machine reading comprehension tasks (the below three figures). The grey and orange lines are the results of THG and Transformer, respectively.

| Model | D=200 | D=300 | D=400 | D=500 |
|---|---|---|---|---|
| Transformer | 36.64% | 38.17% | 37.13% | 37.52% |
| THG | 37.14% | 38.24% | 37.71% | 40.01% |

Table 1: Results of classification task for different embedding dimension (200, 300, 400 or 500).

| Compatibility Function | $F$-score | Time(s) |
|---|---|---|
| dot product (our) | 73.22% | 1026 |
| hyperbolic distance | 61.74% | 15612 |

Table 2: Results for choice of compatibility function on BC5CDR disease with dimension of 300.

## 3.2 Implementation Details

We use randomly initialized word embeddings. The hidden dimension is the same as embedding dimension. The feed-forward dimension is 2 times of the hidden dimension. The numbers of heads and layers are 5 and 1, respectively. For sequence labeling task and classification task, we train our model using Adam with a learning rate of 1e-3 for optimization. For machine reading comprehension task, we apply the Rmsprop optimization algorithm with a learning rate of 5e-4.

## 3.3 Main Result

Figure 3 and Table 1 present comparisons between Transformer and THG. In high dimensions, THG achieves huge gain improvements. And in low dimensions, they perform similarly. We explain this behavior by noting that when the dimension is large, THG have enough capacity to automatically choose task-oriented distribution. When the dimension get higher, the performances of THG keep high, which demonstrates THG could alleviate overfitting.

## 3.4 Choose of Compatibility Function

For compatibility function, we choose dot product, while the common choice in hyperbolic is hyperbolic distance. Table 2 shows that dot product can get better performance with less time than hyperbolic distance, which fits the theory in the Section 2.3.

## 4 Conclusion

We introduce THG, an encoder model that leverages the properties of both Euclidean and Hyperbolic space. THG leverages hyperbolic liner to map query and key into the task-related distribution automatically. For effectiveness and time-efficient, THG adopts dot product. THG gives state-of-the-art performance on a number of tasks such as sequence labeling, machine reading comprehension and classification. Future directions for this work include exploring other hyperbolic models or kernels of SVM that might benefit the distribution choice.